\documentclass[sigconf]{acmart}

\usepackage{balance} 
\usepackage{enumitem}
\usepackage{titlecaps}
\usepackage{mfirstuc}
\usepackage{bm}
\usepackage{enumitem}
\usepackage{wrapfig}
\usepackage{adjustbox}
\usepackage{amsmath}
\usepackage{multicol}
\usepackage{multirow}
\usepackage{subcaption}

\usepackage{amssymb}
\usepackage{amssymb}
\usepackage{booktabs} 
\usepackage{xcolor,colortbl}
\usepackage{float}
\hypersetup{
    colorlinks=true,     
    citecolor=blue,     
    linkcolor=red,       
    urlcolor=blue,      
    filecolor=purple    
}

\usepackage{color}
\definecolor{mygray}{gray}{.9}
\usepackage{colortbl} 
\newcommand{\pub}[1]{{\color{gray}{\tiny{[{#1}]}}}} 
\usepackage{arydshln} 
\usepackage{makecell} 

\usepackage{subcaption}
\usepackage{booktabs}
\usepackage{multirow}
\usepackage{colortbl}

\newcommand{\myheading}[1]{\noindent \textbf{#1}}

\AtBeginDocument{%
  }
\setcopyright{acmlicensed}
\copyrightyear{2026}
\acmYear{2026}
\acmDOI{XXXXXXX.XXXXXXX}
\acmConference[Conference acronym 'XX]{Make sure to enter the correct
  conference title from your rights confirmation email}{June 03--05,
  2018}{Woodstock, NY}
\acmISBN{978-1-4503-XXXX-X/2018/06}

\acmSubmissionID{5053}

\begin{document}
\definecolor{yellow}{rgb}{1, 1, 0.7}
\definecolor{orange}{rgb}{1, 0.85, 0.7}
\definecolor{red1}{rgb}{1, 0.7, 0.7}
\title[Fashion-3DLR.]{
Fashion-3DLR: A Controllable 3D Garment Generation Using Pairwise Fashion Elements for Intelligent Design
}

\author{Shenghao Yang}
\email{2242164@mail.dhu.edu.cn}
\affiliation{%
  \institution{Donghua University}
  \city{Shanghai}
  \country{China}
}

\author{Hongtao Zhang}
\email{2201957@mail.dhu.edu.cn}
\affiliation{%
  \institution{Donghua University}
  \city{Shanghai}
  \country{China}
}

\author{Yuhan Yi}
\email{yuhanyi@mail.dhu.edu.cn}
\affiliation{%
  \institution{Donghua University}
  \city{Shanghai}
  \country{China}
}

\author{Zhihao Tang}
\email{220995117@mail.dhu.edu.cn}
\affiliation{%
  \institution{Donghua University}
  \city{Shanghai}
  \country{China}
}

\author{Zihao Cui}
\email{20995127@mail.dhu.edu.cn}
\affiliation{%
  \institution{Donghua University}
  \city{Shanghai}
  \country{China}
}

\author{Lian Wen}
\email{210910807@mail.dhu.edu.cn}
\affiliation{%
  \institution{Donghua University}
  \city{Shanghai}
  \country{China}
}

\author{Han Yan}
\email{yanhan@dhu.edu.cn}
\affiliation{%
  \institution{Donghua University}
  \city{Shanghai}
  \country{China}
}

\author{Yuan Gao}
\email{gaoyuan@pjlab.org.cn}
\affiliation{%
  \institution{Shanghai Artificial Intelligence Laboratory}
  \city{Shanghai}
  \country{China}
}

\author{Mingbo Zhao}
\authornote{Corresponding author}
\email{mzhao4@dhu.edu.cn}
\affiliation{%
  \institution{Donghua University}
  \city{Shanghai}
  \country{China}
}



\begin{abstract}

AI-generated content (AIGC) has made significant progress, with 2D generative models becoming ready-to-use tools for the digital fashion industry. 
However, 3D garment generation remains in its nascent stage, where in the realm of fashion, the semantic information of diverse design elements exhibits intricate coupling relationships in 3D representations, posing substantial challenges for generating diverse 3D garments.
In this work, to handle the above problem, We introduce Fashion-3DLR, a novel 3D garment generation framework that utilizes diverse design elements to create high-quality, versatile 3D garment assets. 
Specifically, to bridge the semantic gaps between different fashion elements, we propose a Garment Feature Fusion Diffusion Transformer (GFF-DiT) module to integrate 2D fashion design elements, e.g., sketch and texture, into latent space. 
Within the latent space, we then employ a rectified flow transformer to generate geometry latents, which can be decoded into various 3D garment representations, including 3D Gaussians and meshes.
Furthermore, we integrate Fashion-3DLR into downstream tasks, achieving the 3D Gaussian Splatting (3DGS)-driven cloth physical simulation and mesh-based virtual try-on. 
Experimental results indicate that Fashion-3DLR surpass the previous state-of-the-art methods, which verify that the proposed work can generate well-structured, non-watertight garments capable of physical simulation and virtual try-on, underscoring its potential as a versatile 3D garment design tool.

\end{abstract}




\maketitle

\section{Introduction}

Ready-to-wear fashion design necessitates iterative deconstruction and recombination of multi-modal elements, including conceptual sketches, fabric samples, and brush paintings, to transform creative ideas into tangible realities.
Fortunately, the advent of generative models, notably Generative Adversarial Networks (GANs)~\cite{creswell2018generative} and Stable Diffusion (SD)~\cite{rombach2022high}, has precipitated the maturation of neural network (NN)-driven methodologies in digital fashion design.
Integration of generative artificial intelligence (AI) frameworks into the design workflow empowers designers to dynamically modify stylistic and structural garment attributes, thereby lowering the industry's entry barrier for newcomers \cite{bian2025chatgarment}.
Concurrently, foundational advancements in diffusion models have catalyzed progress toward 3D digital garment synthesis \cite{he2024dresscode}.
An effective digital garment creation tool should allow users to customize their outfits in a manner that highlights the individuality and diversity of human beings, with various garment attributes.

Recent breakthroughs in latent diffusion models have significantly advanced the state-of-the-art (SOTA) in digital garment asset generation~\cite{liu2023towards}.
These frameworks enable novice users to iteratively refine fashion designs through multi-modal conditions, including text, sketches, textures, and colors.
Especially, FashionDiff~\cite{yan2023fashiondiff} leverages paired fashion-related elements (e.g., contour, textures, and brush area) to creatively generate editable fashion item images by learning from human designers' styles.
Compared to 2D garments, 3D garments exhibit superior geometric fidelity and richer structural information. 
Despite growing industry demand for 3D garments, the field of 3D garment editing remains in the nascent stages of development.


\begin{figure*}[t]
  \centering
  \includegraphics[width=0.75\linewidth]{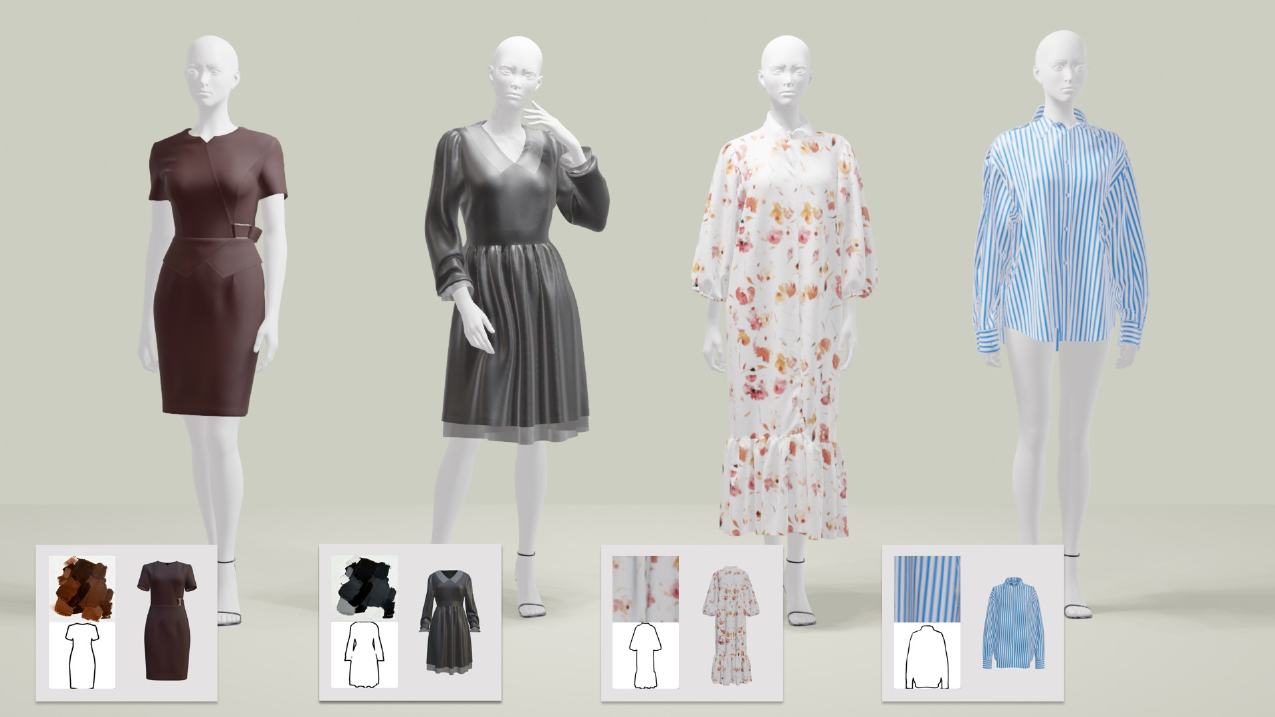}
       \caption{ 
Given paired design elements, e.g., sketch and texture, Fashion‑3DLR generates a diverse set of high‑quality, wearable 3D garments. 
Fashion‑3DLR synthesizes richly detailed 3D models by preserving the structural content encoded in the sketch while integrating material characteristics conveyed by the texture or brush strokes, thereby producing garments with faithful and realistic fabric appearance.
       }
   \label{fig:teaser}
\end{figure*}

Developing a controllable 3D fashion generation model capable of simultaneously processing multi-modal conditions presents three fundamental challenges.
(1) \textbf{Cross-modal semantic disambiguation}. 
The model must effectively bridge the substantial semantic gap between sketch-based structural representations and appearance-based auxiliary features, such as textures and colors~\cite{liu2023towards, he2024dresscode, korosteleva2022neuraltailor}. 
Furthermore, preserving the unique characteristics of both elements within a highly integrated 3D representation is hard-handling.
(2) \textbf{High-quality generation}.
Current 3D generation architectures~\cite{long2024wonder3d,wang2021neus,tang2023dreamgaussian,liu2024one,liu2023syncdreamer} often suffer from significant geometric detail loss during both the feature extraction and decoding phases. 
To provide users with realistic fashion designs, the model should generation high-quality 3D models with precise texture details.
(3) \textbf{Intuitive edit-ability}.
The traditional paradigms of 3D modeling, which rely on specialized expertise and intricate manual parameter tuning, is both time-consuming and technically prohibitive~\cite{achiam2023gpt}.
Thus, our solution requires a user-friendly, interactive editing workflow, as intuitive as chatting with an LLM-based AI agent.

Current AIGC systems still face fundamental limitations in performing high-level creative tasks such as 3D garment synthesis. 
Integrating heterogeneous design elements, maintaining fine-grained details under sparse training data, and achieving physically plausible simulations remain major challenges.
We introduce a novel 3D garment generation framework, i.e., Fashion-3DLR, that simultaneously addresses these limitations through cross-modal fusion of multi-semantic inputs, detail enhancement via 2D priors, and physics-aware Gaussian kernel deformation. 
Fashion-3DLR establishes the first end-to-end pipeline capable of generating high-fidelity 3D garments with dynamic simulation properties while maintaining precise artistic control, significantly reducing the manual refinement typically required in professional design workflows.
Our key contributions are as follows:
\begin{itemize}
\item We propose Fashion‑3DLR, a unified framework that synthesizes high‑quality 3D garments from paired design elements by integrating sketch‑driven structural modeling with texture‑guided appearance generation. 
The framework unifies cross‑modal semantic fusion, rectified‑flow‑based latent generation, and multi‑format decoding, enabling high‑fidelity outputs in both 3DGS and mesh.
\item We introduce GFF‑DiT, a diffusion‑based cross‑modal fusion module that jointly encodes sketch geometry and texture semantics through bidirectional modulation and multi‑level feature aggregation. 
This design preserves structural fidelity while enhancing fabric‑level details, enabling diverse and high quality 3D garment synthesis under limited 3D supervision.
\item Fashion-3DLR can be integrated with 3D Gaussian-based physical simulation, where Gaussians are modeled as deformable particles to enable realistic cloth dynamics.
We enable realistic, fabric‑specific dynamics and support the simulation of diverse materials, including cotton, silk, wool, and nylon, while preserving high‑frequency visual details.
\item Fashion‑3DLR supports mesh decoding for downstream virtual try‑on, producing garments with accurate geometry, consistent textures, and strong alignment with input sketches and textures. 
Compared with existing sewing pattern-based methods, Fashion‑3DLR achieves superior realism, structural fidelity, and user‑preferred try‑on quality.
\end{itemize}

\begin{figure*}[t]
  \centering
  \includegraphics[width=1\linewidth]{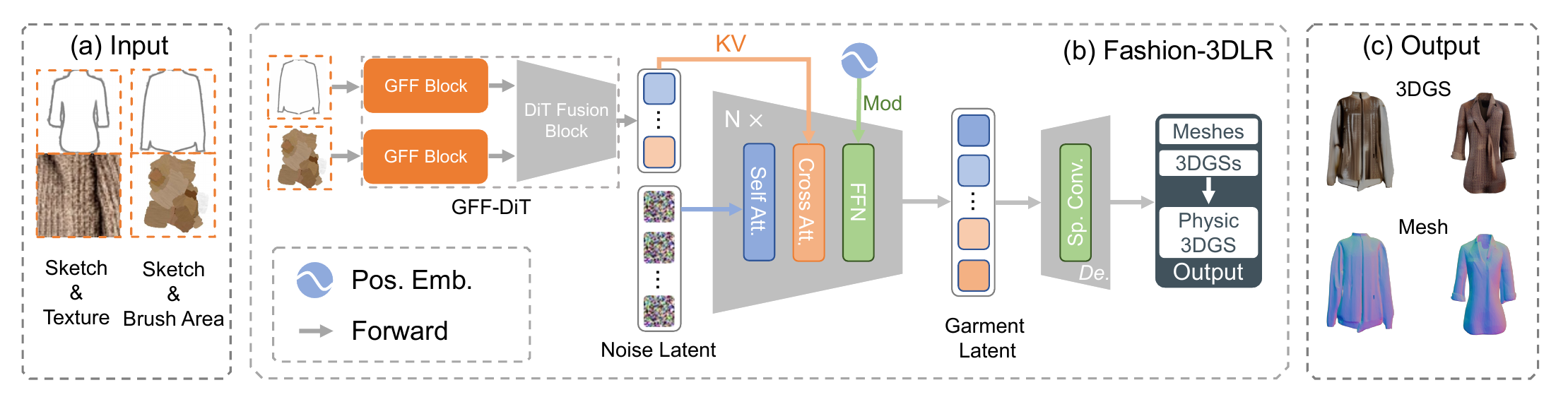}
       \caption{ Overall pipeline for Fashion-3DLR. Our approach processes paired design element images by fusing their features through GFF-DiT. 
       These features then interact via multi-head attention mechanisms and noisy latents in our custom rectified flow transformers. 
       This interaction produces shape latents with integrated semantic features, which a VAE decoder transforms into high-quality 3D clothing assets. 
       }
   \label{fig:overview}
\end{figure*}

\section{Related Work}
\myheading{Heterogeneous Semantic Fusion Generation.}
In the fields 2D and 3D generation, researches on fusing inputs that share the same data type but contain different semantic elements continues to advance.
In the 2D domain, methods~\cite{cui2018fashiongan,dong2020fashion,jiang2021deep,nichol2021glide,zhu2017your} address unpaired data translation through specialized loss functions or shared latent spaces. 
Specifically, FashionDiff~\cite{yan2023fashiondiff} focuses on fashion image generation by creating paired data such as sketches and textures, utilizing the attention-based fusion process to eliminate semantic differences and merge them into the latent space, combined with the ControlNet-style generator to achieve flexible control over multiple elements~\cite{zhang2023adding,rombach2022high}.
In the 3D domain, excellent 3D generation works such as SDFusion~\cite{cheng2023sdfusion}, Trellis~\cite{xiang2025structured}, and CraftsMan3D~\cite{li2024craftsman3d} not only ensure high-quality generation results but also possess good text comprehension abilities, capable of effectively balancing different semantics within prompts. 
However, their exploration of heterogeneous semantic understanding for image inputs remains preliminary, only supporting multi-image inputs with strong consistency, such as multi-view images of the same object~\cite{long2024wonder3d,wang2021neus,tang2023dreamgaussian,liu2024one,liu2023syncdreamer}.
In this paper, Fashion-3DLR achieves 3D generation through heterogeneous semantic image fusion.
It can simultaneously learn and preserve features from both sketch images and design element images that have no consistency between them to generate 3D clothing assets.

\myheading{2D Generative Models for 3D Creation.}
The exceptional generalization capabilities of 2D generative models~\cite{shi2023zero123++, liu2023zero, liu2023syncdreamer,chen2024cascade} have sparked considerable interest in research exploring their application to 3D asset creation~\cite{long2024wonder3d,wang2021neus,tang2023dreamgaussian,liu2024one,liu2023syncdreamer} in recent years.
DreamFusion~\cite{poole2022dreamfusion} established the foundation for this direction by distilling knowledge from pretrained image diffusion models to optimize 3D assets. 
Subsequent works further improved 3D generation quality through more advanced distillation techniques, such as improved score distillation sampling, introduction of multi-view consistency constraints.
However, 2D generative models inherently lack the ability to model three-dimensional spatial consistency. 
This limitation leads to perspective contradictions in the generated multi-view images. 
As a result, 3D assets reconstructed using these methods generally exhibit poorer geometric accuracy and less detailed features compared to native 3D generative models~\cite{skorokhodov2022epigraf,liu2024one} that learn directly from 3D datasets.
Meanwhile, Trellis~\cite{xiang2025structured} introduced an effective visual feature aggregation method that maps image features to the 3D shape latent space, enhancing both structural integrity and material detail in the generated results.
Drawing inspiration from this approach, we designed rectified flow transformers that effectively incorporate GFF-DiT fused image features into the 3D generation process.


\myheading{3D Gaussian-Based Cloth Simulation.}
Recent works have explored incorporating 3D Gaussian Splatting~\cite{kerbl20233d} into garment simulation pipelines.
However, most existing methods~\cite{rong2024gaussian,li2025garmentdreamer,sarafianos2024garment3dgen} remain fundamentally mesh-based.
In these approaches, cloth dynamics are first simulated using conventional mesh-based physical models, after which 3D Gaussians are applied as a texture or appearance layer mapped onto the mesh surface.
Such hybrid designs largely inherit the limitations of traditional graphics pipelines and introduce additional artifacts during the mapping process, often leading to loss of fine-grained details and rendering quality that is inferior to native mesh-based materials.
To move beyond mesh-dependent formulations, PhysGaussians~\cite{xie2024physgaussian} proposed a preliminary attempt at directly simulating 3D Gaussians by leveraging the Material Point Method (MPM)~\cite{hu2018moving}, where 3D Gaussian kernels are interpreted as particle primitives to enable physically plausible motion.
MPM is a hybrid simulation framework that combines Lagrangian particles with Eulerian grids, and has demonstrated strong capability in handling large deformations, topological changes, and complex frictional interactions.
It has been widely applied to multi-physics scenarios involving elastic materials, fluids, and granular media such as sand and snow~\cite{jiang2015affine,klar2016drucker,stomakhin2013material}.
Building upon these insights, we further advance this direction by developing a fully 3D Gaussian-based cloth simulation framework.
Unlike prior mesh-dependent approaches, our method eliminates the need for intermediate mesh representations and operates directly on 3D Gaussian primitives.
This design preserves the rich visual fidelity of 3D Gaussians assets while enabling realistic cloth dynamics, thereby expanding the applicability of 3D Gaussians for garment modeling and simulation.
\section{Methods}
\label{sec:Methods}

\myheading{Overview.}
The Fashion-3DLR framework introduces a novel approach for controllable 3D garment generation by leveraging pairwise fashion elements, such as sketches and textures, to create high-quality and versatile 3D garment assets. 
The framework addresses three key challenges in 3D fashion generation: cross-modal semantic disambiguation, high-quality generation, and intuitive edit-ability. 
The pipeline begins with the Garment Feature Fusion Diffusion Transformer (GFF-DiT), which extracts and fuses semantic information from heterogeneous design elements (e.g., sketches and textures) into a unified latent space. 
This fusion is achieved through a bidirectional modulation process that ensures the preservation of distinctive features from each input modality. 
The fused features are then passed to a rectified flow transformer, which generates garment latents with integrated 3D structural and appearance details. 
These latents are subsequently decoded into multiple output formats, including 3DGS representations and meshes, enabling flexible downstream applications.
 


\begin{figure}[t]
    \centering
    \includegraphics[width=1\linewidth]{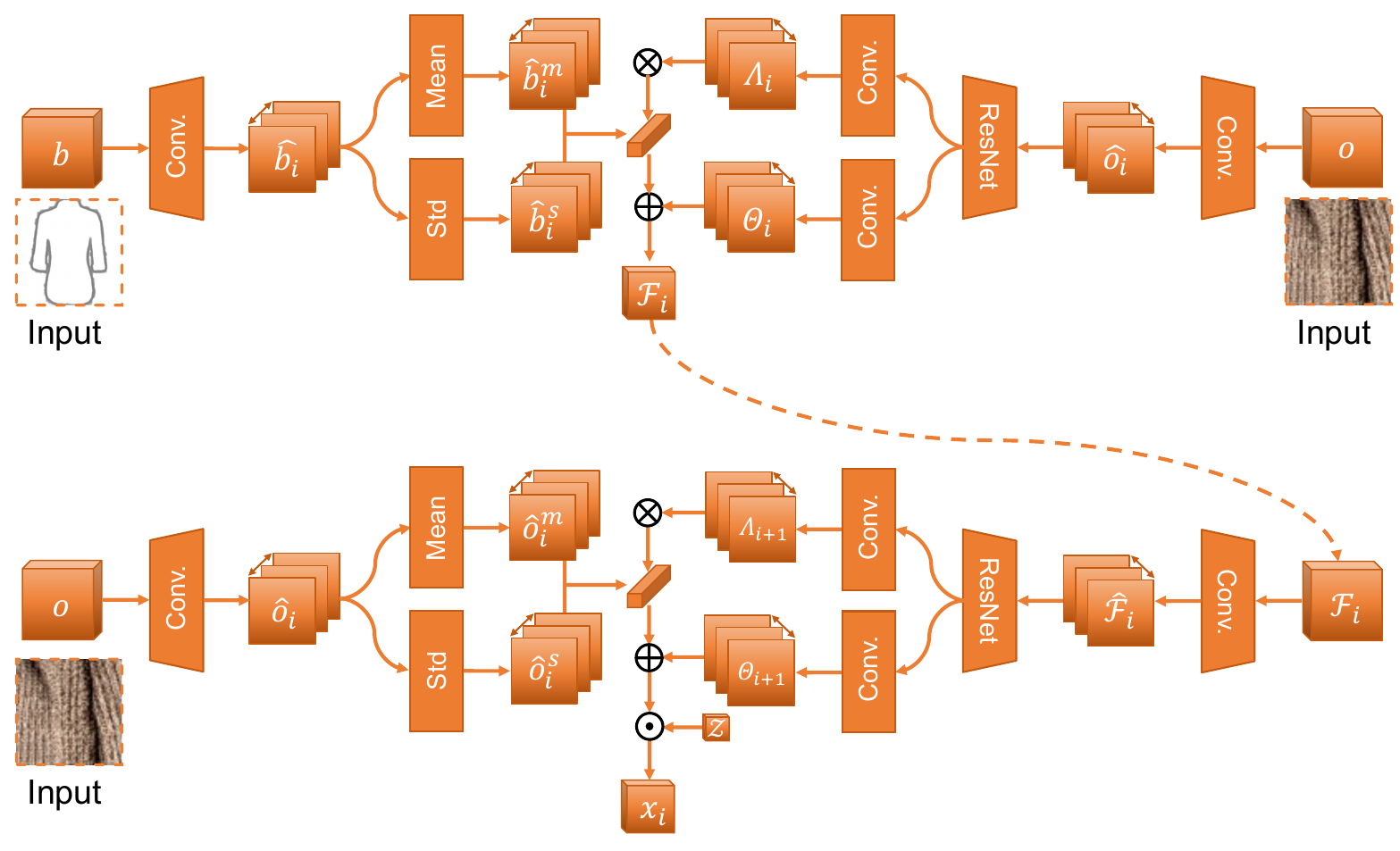}
    \caption{Details of the proposed GFF process.}
    \label{fig:GFF}
\end{figure}

\subsection{Garment Feature Fusion-based DiT}
The Garment Feature Fusion (GFF) serves as the feature extract and fusion module within Fashion-3DLR framework.
Fig.~\ref{fig:GFF} presents the proposed GFF module, which begins with the extraction of semantic information from two conditions, i.e., sketch and texture, and then these elements are seamlessly integrated into latent spaces~\cite{yan2023fashiondiff}.
Especially, $b$ and $o$ are fed into convolutional layers for down-sampling to gain the intermediate variable $\hat{b}_i$ and $\hat{o}_i$, respectively. 
Subsequently, $\hat{o}_i$ is first fed into a residual block and then passed through two different convolutional layers to produce two learnable factors $\Lambda_{i}$ and $\Theta_{i}$, denoted as $\mathcal{G}(o) \rightarrow\left\{\Lambda_{i}, \Theta_{i}\right\}$.
The extracted sketch feature $\hat{b}_i$ is convolved to obtain the mean and variance, which is calculated as $\hat{b}_i^m={mean}( \hat{b}_i )$ and $\hat{b}_i^s={var}( \hat{b}_i )$, respectively. 
To enable interactive communication between the sketch and texture, we modulate the sketch feature $\hat{b}_i$ with two learnable factors $\Lambda_{i}$ and $\Theta_{i}$ determined by feature $\hat{o}_i$, denoted as:
\begin{equation}
\mathcal{F}_{\left\{\Lambda_i, \Theta_i\right\}}(b)=\left(\Theta_i \mid \mathcal{G}({o})\right) \otimes\left[\frac{\hat{b}_i-\hat{b}_i^m}{\hat{b}_i^s}\right] \oplus\left(\Lambda_i \mid \mathcal{G}(o)\right)
\end{equation}
where $\oplus$ denotes multiplication and $\otimes$ indicates element-wise addition. 
To further facilitate bi-directional interaction between the sketch and texture conditions, we perform $\mathcal{G}((\mathcal{F}_{\left\{\Lambda_i, \Theta_i\right\}}(b))) \to \left\{\Lambda_{i+1}, \Theta_{i+1}\right\}$, thus we have:
\begin{equation}
\mathcal{F}_{\left\{\Lambda_{i+1}, \Theta_{i+1}\right\}}(o)=\Theta_{i+1} \otimes\left[\frac{\hat{o}_i-\hat{o}_i^m}{\hat{o}_i^s}\right] \oplus\Lambda_{i+1}
\end{equation}
Finally, $\mathcal{F}_{\left\{\Lambda_{i+1}, \Theta_{i+1}\right\}}(o)$ is passed to a zero convolution layer $\mathcal{Z}$ to obtain the fusion feature $x_i$, denoted as: 
\begin{equation}
x_i=\mathcal{Z} \bullet \mathcal{F}_{\left\{\Lambda_{i+1}, \Theta_{i+1}\right\}}(o)
\end{equation}
It is important to highlight that the symmetrical treatment of $b$ and $o$, which has the potential to eliminate disparities within the data, ultimately leading to a robust and accurate representation of visual semantics.
In a similar manner, we proceed with the aforementioned steps to derive the values of $x_{i+1}$, $x_{i+2}$, $x_{i+3}$, denoted $\mathcal{X}_{i}=\{x_{i}, x_{i+1}, x_{i+2}, x_{i+3} \}$.

In the subsequent DiT fusion block, the input condition $\mathcal{X}_{i}$ is first transformed through a linear layer to produce latent noise. 
This latent noise, along with the time step embedding, is then processed by the DiT module~\cite{peebles2023scalable} to generate fused conditional features $\mathcal{M}$. 
These features subsequently serve as the controlling condition for the 3D garment generation.

\subsection{Garment Latents Generation}
We introduce a rectified flow transformer (RFT) for generating garment latents $\mathcal{S}$, as shown in Fig.~\ref{fig:overview}. 
The transformer block is sequentially composed of a self-attention layer, a cross-attention layer, and a feed-forward network. 
An input dense noisy grid is serialized into the noise latent, combined with positional encodings, and fed into the transformer blocks for denoising. 
We divide 3D garment assets into voxel grids, where each grid contains activated voxel markers and visual features that capture the detailed structure and appearance of the local area.
Specifically, we render images by randomly sampling camera perspectives on a sphere, extract feature maps using pretrained encoders~\cite{oquab2023dinov2}, project each voxel onto these multi-view feature maps to retrieve features from corresponding positions, and then average these features to create the final visual feature~\cite{xiang2025structured}.
We align the resolution of voxelized features with garment latents, enabling transformer blocks to effectively denoise by combining visual features with both the strong representational capabilities of relevant features and the structural framework in activated voxels.

Condition $\mathcal{M}$ is injected through cross-attention layers as keys and values, and the noise latent is injected through the self-attention layer as a query. 
After passing through $N$ transformer blocks, we obtain a garment latent $\mathcal{S}$ which not only reflects the input texture $o$ and sketch $b$ but also possesses the correct 3D features. 
Different convolutional up-sampling blocks are appended at the end of the transformer blocks to convert the garment latent $\mathcal{S}$ into high-quality 3D assets in various formats, including 3DGSs and meshes.

\subsection{Optimization of Fashion-3DLR}
We employ a two-stage training approach to optimize our proposed Fashion-3DLR framework. 
For the GFF-DiT module, in addition to the initial image $z_0$ and the noise image $z_t$ generated at time $t$, we incorporate fashion-related inputs including fashion-specific conditions (e.g., sketch $b$ and texture $o$) and the current time step $t$.
We train the GFF-DiT as follows: 
\begin{equation}
\mathcal{L}^{G F F}(\theta)=\mathbb{E}_{z_0, t, o, b, \epsilon}\left[\left\|\epsilon-\epsilon_\theta\left(z_t, t, \mathrm{~f}(o, b)\right)\right\|_2^2\right]
\end{equation}
where $\mathrm{~f}$ represents the proposed GFF process and $\epsilon_\theta$ is the DiT network. 
Upon completion of training, we freeze the relevant network weights of GFF-DiT. 
For the rectified flow transformer module, we utilize a linear interpolation forward process. 
Specifically, given data samples $n_0$ and noises $\varepsilon$ with a timestep $t$, the forward and backward process is denoted as $n(t)=(1-t)n_{0}+t\varepsilon$ and ${v}(n, t)=\nabla_t n$, respectively. 
Consequently, the rectified flow transformer ${v}_\theta$ can be approximated by minimizing the conditional flow matching (CFM) objective, denoted as:
\begin{equation}
\mathcal{L}^{C F M}(\theta)=\mathbb{E}_{t, n_0, \varepsilon}\left[\left\|{v}_\theta(n, t)-\left(\varepsilon-n_0\right)\right\|_2^2\right]
\end{equation}
After training, the garment latents $\mathcal{S}$ can be generated by the Fashion-3DLR and converted into high-quality 3D assets.

\begin{figure*}[htbp]
  \centering
  \includegraphics[width=0.98\linewidth]{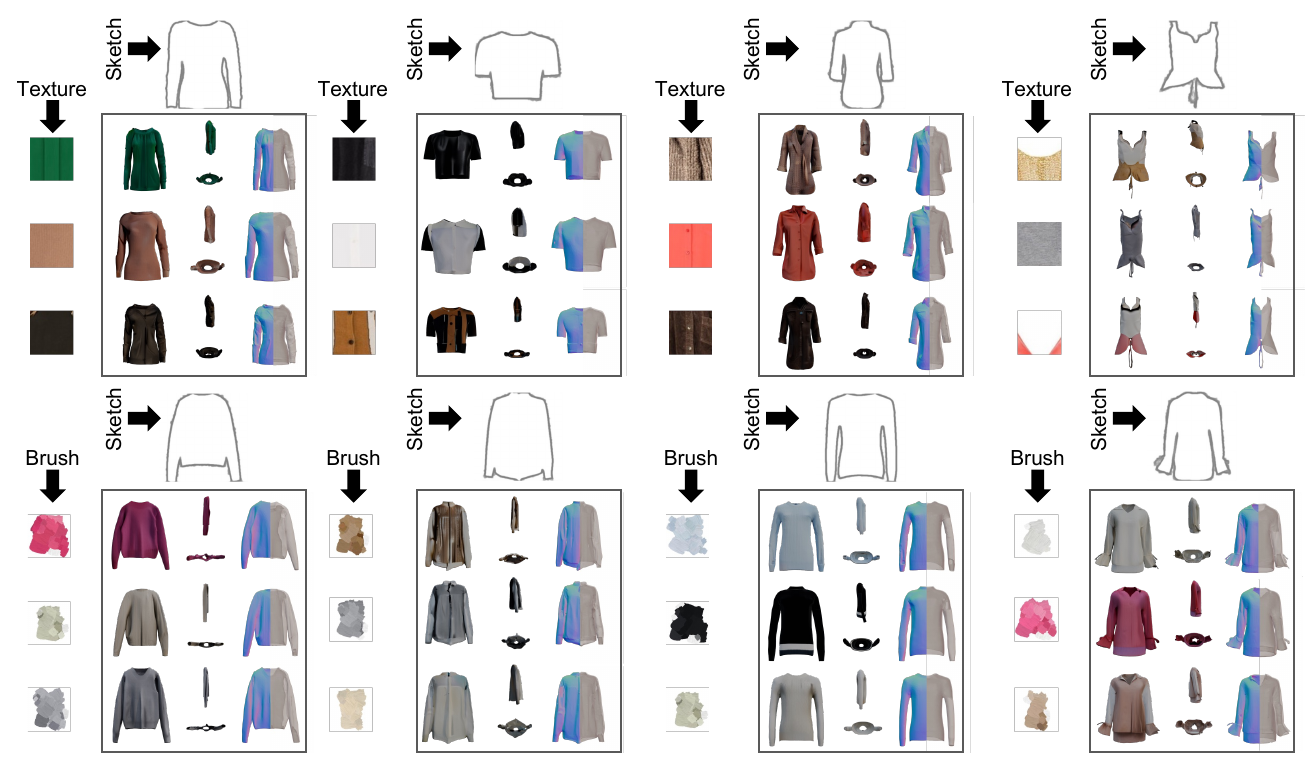}
       \caption{Qualitative results using sketch images with texture images, and sketch images with oil brush images.}
   \label{fig:generation}
\end{figure*}

\subsection{Physical Garment 3DGS}

In this section, we highlighted that Fashion-3DLR supports 3DGS-based garment asset generation. 
To estimate fabric‑specific physical parameters for cotton, silk, wool, and nylon, we adopt a physics integrated 3D Gaussian representation inspired by PhysGaussian. 
A static cloth instance is first reconstructed as a set of unstructured 3D Gaussian kernels $\{x_p,\ \sigma_p,\ A_p,\ C_p\}_{p \in \mathcal{P}} $, 
where $x_p$, $\sigma_p$, $A_p$, $C_p$ denote the Gaussian centers, opacities, covariance matrices, and spherical harmonic coefficients. 
For each fabric type, we instantiate a corresponding material model from a fabric library and generate its ground‑truth dynamic behavior by simulating wind‑driven cloth motion in a commercial cloth simulator~\cite{bouman2013estimating}.
We then construct a 3D‑Gaussian cloth with identical initial geometry and endow each Gaussian with continuum‑mechanics‑based kinematics using the MPM solver. 
During parameter identification, we iteratively adjust a compact set of physically meaningful parameters $\vartheta = \{k_b,\ k_s,\ k_{sh},\ \rho\}$, 
where $k_b$ refers to bending stiffness, $k_s$ refers to stretching stiffness, $k_{sh}$ refers to shear stiffness, and $\rho$ refers to area density. 
These parameters govern the evolution of the Gaussian field through
\begin{equation}
A_p(t) = F_p(t), A_p, F_p(t)^{\top}, \qquad x_p(t) = \phi(X_p, t)
\end{equation}
where $F_p(t)$ and $\phi(\cdot)$ denote the deformation gradient and deformation map. 
We optimize by minimizing the discrepancy between the simulated Gaussian dynamics and the reference cloth video:
\begin{equation}
\mathcal{L}^{\text{gauss}}(\vartheta) = \arg\min_{\vartheta} \left|\left|F^{\text{video}} - F^{\text{gauss}}(\vartheta) \right|\right|^{2}
\end{equation}
where $F^{\text{video}}$ contains motion features extracted from the baseline cloth video, e.g., local vibration frequencies, modal responses, and motion‑energy spectra, and $F^{\text{gauss}}(\vartheta)$ denotes the same features computed from the physics‑driven Gaussian evolution. 
Through the above process, we enables physically consistent and fabric‑specific parameter identification directly within the 3D Gaussian domain.

\section{Experiments}

In this section, we conduct both quantitative and qualitative comparative experiments to evaluate artistic 3D generation by integrating design elements. 
We include general 3D generation models supporting multi-image inputs and existing clothing generation methods based on sewing patterns.

\begin{figure*}[htbp]
  \centering
  \includegraphics[width=1\linewidth]{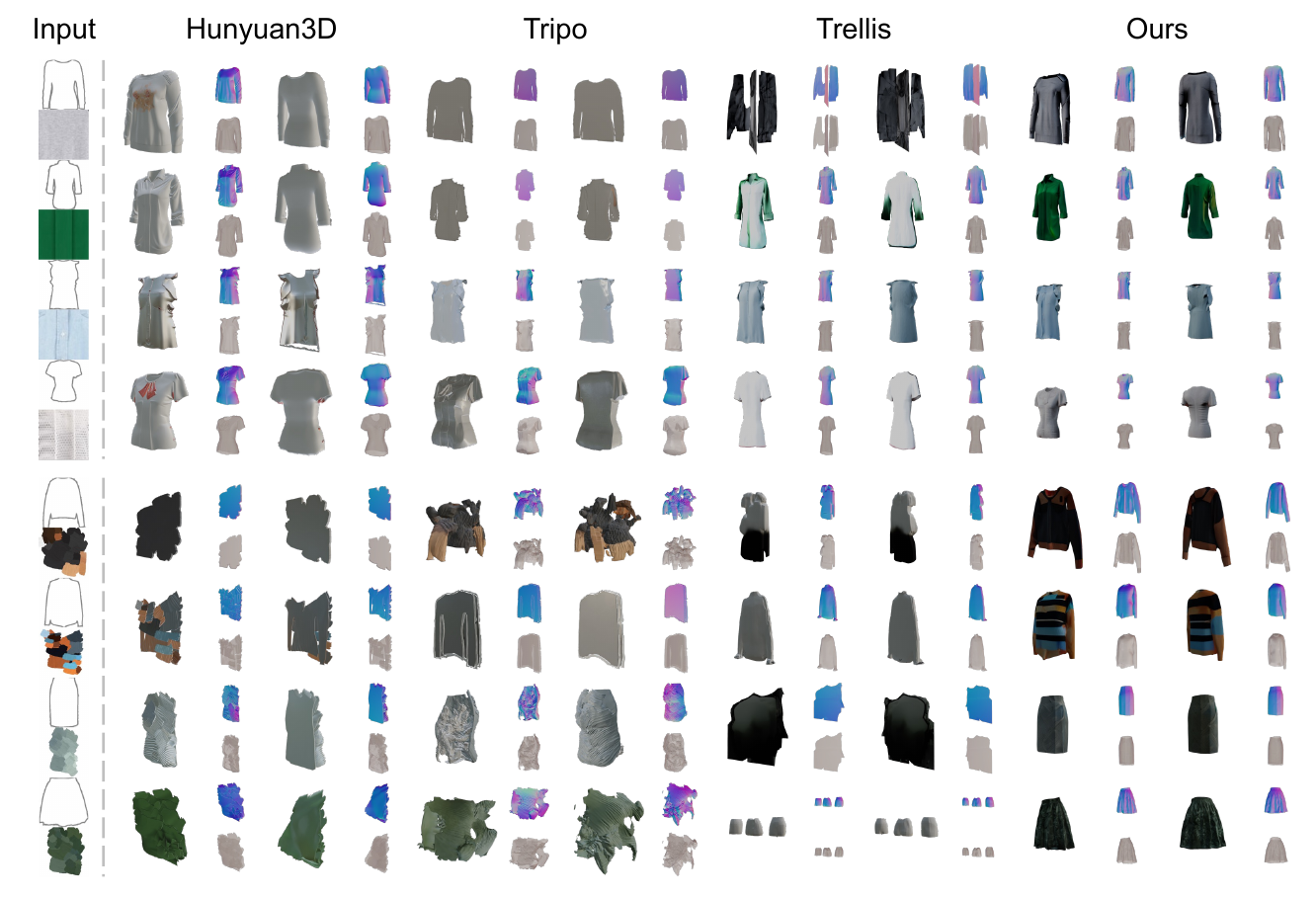}
       \caption{Qualitative comparisons with baselines for 3D garment generation with fusion of sketch, texture and oil brush.}
   \label{fig:qualitative}
\end{figure*}

\subsection{Experiment Setup}

\myheading{Dataset.}
For training, we fine-tuned the pre-trained RFT structure on two datasets: the SewFactory dataset~\cite{liu2023towards} containing 19.1k high-quality 3D garment assets, and the Polyvore fashion image collection~\cite{yan2023fashiondiff} comprising 92,488 items across 6 categories. 
For SewFactory, we rendered 120 multi-view images per 3D asset and used GPT-4o~\cite{achiam2023gpt} to generate descriptive captions for both rendered and Polyvore images. 
To construct multi-modal supervision, we applied Holistically-Nested Edge Detection (HED)~\cite{xie2015holistically} to obtain sketch images, employed neural painting to synthesize oil-brush stylization, and adopted 32×32 random foreground window sampling to extract texture patches. 
This pipeline yielded paired instance–element data, which were further augmented in text, image, and 3D modalities, with all images standardized to a 256×256 resolution.


\myheading{Metrics.} 
We introduced the following metrics to evaluate Fashion-3DLR.
\myheading{FID$_{\text{CLIP}}$} measures the semantic consistency between the generated garments and the input texture or brush by computing the distance between their CLIP‑based feature distributions, where a higher score indicates better alignment in color and details.
\myheading{LPIPS} evaluates the contour similarity between generated garments and sketch images by comparing deep feature distances, with lower values indicating closer alignment in shape and structural layout.
\myheading{CLIP-I$_{\text{score}}$} quantifies the semantic similarity between the generated garments and the paired fashion elements by computing their feature alignment in the CLIP image–image embedding space, where a higher score indicates stronger semantic correspondence.

\myheading{Implementation details.}
We first trained GFF-DiT on large-scale image data to strengthen its ability to fuse semantically diverse visual inputs.
The training dataset contains approximately 15.2K samples, with a total training duration of about 30 hours and a single inference time of approximately 40 seconds.
The trained model was then frozen and integrated into a RFT structure for subsequent training on paired 3D instance–element data. 
Detailed training configurations are provided in the appendix.

\begin{table}[htbp]
\centering
\vspace{-10pt}
\caption{
Quantitative comparisons with baseline methods for the task of 3D generation with fusion of sketch images and texture images.
\textbf{Bold} indicates best results.
}
\resizebox{0.48\textwidth}{!}{
\begin{tabular}{lcccc}
\toprule
\rowcolor{gray!20}
\textbf{Method} & \textbf{FID$_{\text{CLIP}}$ $\downarrow$} & \textbf{LPIPS $\downarrow$} & \textbf{CLIP-I$_{\text{score}}$ $\uparrow$} \\
\midrule
Hunyuan3D-3.0~\cite{cao2025hunyuanimage}~\pub{2025}  
& 0.4648 & 0.7001 & 0.7946 \\
Tripo~\cite{TripoSR2024}~\pub{2024}      
& 0.4524 & 0.6803 & 0.7961 \\
Trellis~\cite{xiang2025structured}~\pub{CVPR2025}    
& 0.7142 & 0.7842 & 0.6615 \\
Sparc3D~\cite{lisparc3d}~\pub{NeurIPS2025}    
& 0.8308 & 0.7825 & 0.6638 \\
ReconViaGen~\cite{chang2025reconviagen}~\pub{ICLR2026}    
& 0.8637 & 0.8113 & 0.6395 \\
\rowcolor{gray!20}
Fashion-3DLR~\pub{ours}   & \textbf{0.2743} & \textbf{0.5946} & \textbf{0.9271} \\
\bottomrule
\end{tabular}}
\label{tab:comparisons}
\vspace{-10pt}
\end{table}

\subsection{Controllable 3D Garment Generation}

We evaluate the 3D garment generation quality of Fashion-3DLR through both qualitative analysis and comparison with baseline methods.
We first present representative results produced by our Fashion-3DLR.


\begin{figure}[htbp]
  \centering
  \includegraphics[width=0.90\linewidth]{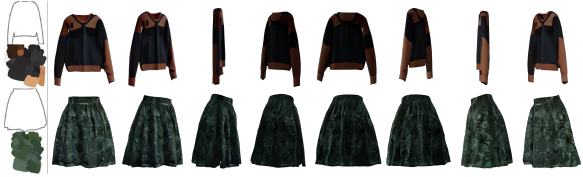}
     \vspace{-10pt}
       \caption{360° demonstration of qualitative results using sketch images with oil brush images.}
   \label{fig:360}
   \vspace{-15pt}
\end{figure}

\myheading{Results Conditioning on Sketch, Texture and Brush Area.}
As illustrated in Fig.~\ref{fig:generation}, Fashion-3DLR effectively fuses input sketches, textures, and artistic oil-brush features to synthesize 3D garments exhibiting diverse and visually coherent styles.
In terms of feature fusion, Fashion-3DLR demonstrates strong capability in translating 2D visual cues into consistent 3D representations in two primary aspects.
First, it adapts minimalist sketch inputs into detailed 3D assets enriched with high-fidelity color textures.
Second, it incorporates brush-stroke-based artistic cues to generate distinctive blending and textural effects across garment surfaces. 
The resulting structured silhouettes align precisely with texture pattern distributions and color gradients, while visible brush traces impart a handcrafted aesthetic that mitigates the uniformity of conventional digital garment.

\myheading{Details Conditioning on Sketch, Texture and Brush Area.}
Regarding shape generation, Fashion-3DLR faithfully preserves both the stylistic intent and geometric outline of the sketch while introducing refined local details. 
It is capable of generating complex band-like structures with bodysuit-level articulation, and perspective renderings reveal non-closed topologies that ensure garment wear-ability. 
These include realistic openings such as circular necklines, ring-shaped cuffs, and open hems, as well as internal cavities that conform to the human body.
Collectively, the above results demonstrate the robustness and versatility of Fashion-3DLR in multi-feature fusion and the creation of high-quality, artistically expressive 3D garment assets.

\subsection{Comparison with SOTA Methods}

To comprehensively evaluate the effectiveness of our proposed Fashion-3DLR framework, we compared Fashion-3DLR with several open-sourced state-of-the-art 3D generation methods.
The qualitative results are presented in Fig.~\ref{fig:qualitative} and Fig.~\ref{fig:360}.

\myheading{Qualitative Comparison for Sketch \& Texture.}
When conditioned jointly on a texture map and a sketch, Fashion-3DLR achieves high-quality fusion of these heterogeneous elements and reconstructs 3D garments with structurally suitable, non-watertight topology. 
Competing methods, however, exhibit notable shortcomings.
\myheading{Hunyuan3D} occasionally recovers valid non-watertight forms but frequently misinterprets rare or atypical sketch geometries, producing front-only shell artifacts. 
Its heavy reliance on sketch background statistics makes it particularly vulnerable to white-pixel contamination, which often washes out the final texture and destroys color consistency.
\myheading{Tripo} shows more severe geometric instability, yielding implausible outputs such as collapsed pancake structures, hollow ring-like artifacts, or even watertight meshes that contradict the intended garment topology.
Its texture fusion remains limited and is similarly degraded by white-background interference.
\begin{figure}[htbp]
  \centering
  \includegraphics[width=0.85\linewidth]{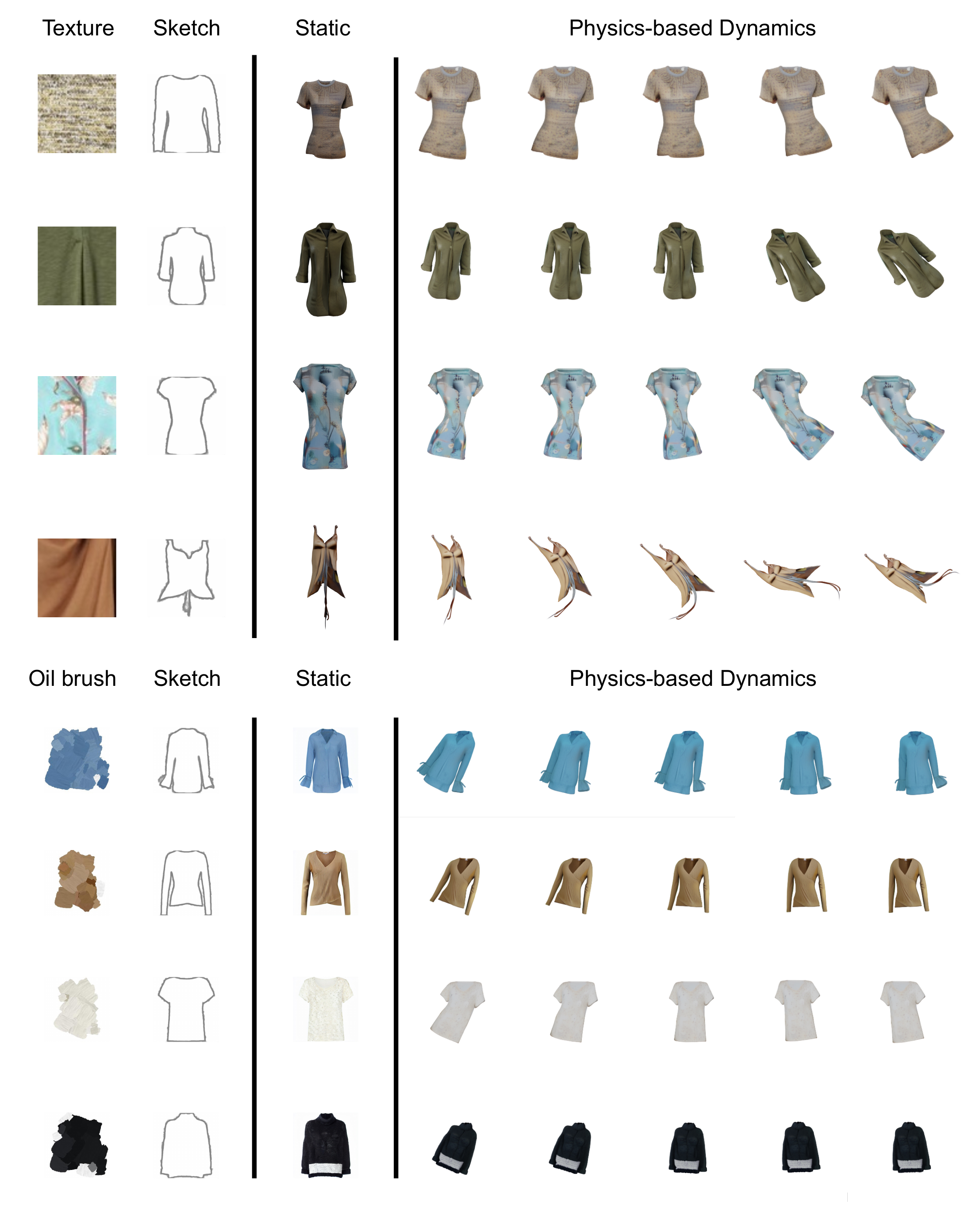}
     \vspace{-15pt}
       \caption{
       Physical simulation demonstration of generated 3DGS clothing assets.
       }
   \label{fig:physics}
   \vspace{-15pt}
\end{figure}
\myheading{Trellis} can sometimes preserve the input sketch but still suffers from unstable reconstruction with frequent topology collapse. 
Although it partially inherits texture colors, most generated surfaces become desaturated or entirely white, with substantial loss of fine-scale detail.
In contrast, \myheading{Fashion-3DLR} consistently reconstructs sketch-aligned, non-watertight garments while preserving high-fidelity texture appearance, demonstrating clear advantages in geometric reliability and texture-geometry consistency.

\myheading{Qualitative Comparison for Sketch \& Oil Brush.}
Under the more challenging setting of cross-domain conditioning with oil-brush artistic inputs and sketch constraints, Fashion-3DLR again exhibits remarkable robustness. 
\myheading{Fashion-3DLR} not only recovers the garment 3D structure dictated by the sketch but also generalizes the brush-stroke color distribution and stylistic cues into the final 3D asset, achieving coherent cross-modal style transfer.
Because the oil-brush areas and sketch contain inherent domain discrepancies, baseline models fail to satisfy either modality. 
\myheading{Hunyuan3D} simply overlays oil-brush areas onto the sketch without structural reasoning, while \myheading{Tripo} and \myheading{Trellis} are unable to produce valid garment shapes altogether.
Qualitative comparisons demonstrate that Fashion-3DLR uniquely maintains structural fidelity, style consistency, and robustness across heterogeneous design inputs, significantly outperforming existing approaches on this complex multi-feature fusion task.

\myheading{Quantitative Comparison.}
We quantitatively evaluate the generation quality based on the rendered results of the final 3D garments, as summarized in Tab.~\ref{tab:comparisons}.
Under the instance–element paired test set derived from SewFactory, Fashion-3DLR consistently surpasses all baseline methods, demonstrating superior performance in both fine-grained detail fidelity and conditional consistency for high-quality 3D garment generation.

\myheading{User Study.}
We conducted a user study, as shown in Tab.~\ref{tab:user-try-on}. 
Participants scored the 3D models generated by each method based on two criteria: overall generation quality and consistency with the input images.
A total of 24 participants were invited to evaluate 200 sets of generated results.

\begin{figure}[htbp]
  \centering
  \includegraphics[width=1\linewidth]{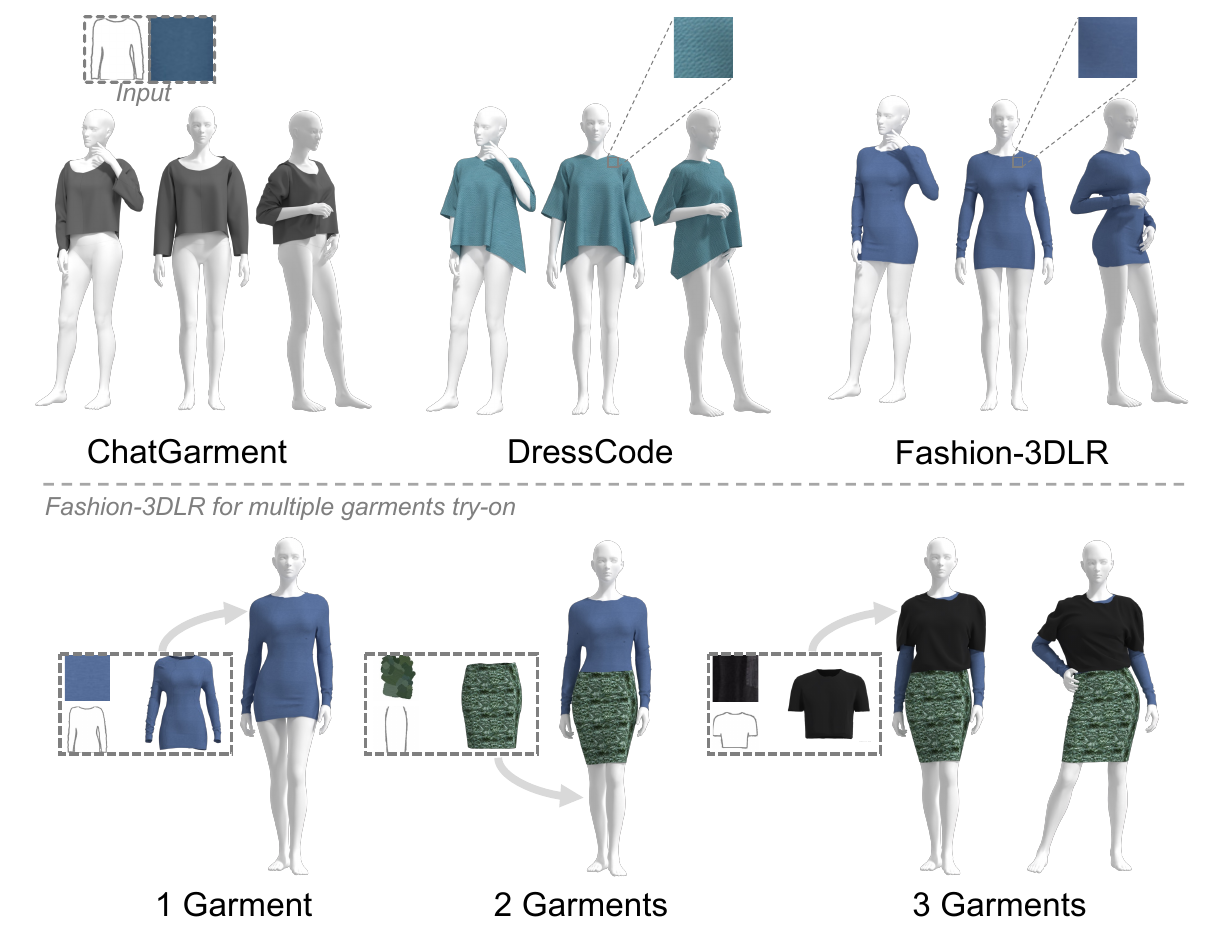}
      \vspace{-10pt}
       \caption{
       Qualitative comparison of existing SOTA clothing generation methods based on sewing patterns with garments draped on human models.
       }
   \label{fig:try-on}
\end{figure}

\begin{table}[htbp]
\centering
\vspace{-10pt}
\caption{
Quantitative comparisons with baseline methods for the task of clothing generation on sewing patterns with garments draped on human models.
\textbf{Bold} indicates best results.
}
\resizebox{0.48\textwidth}{!}{
\begin{tabular}{lcccc}
\toprule
\rowcolor{gray!20}
\textbf{Method} & \textbf{FID$_{\text{CLIP}}$ $\downarrow$} & \textbf{LPIPS $\downarrow$} & \textbf{CLIP-I$_{\text{score}}$ $\uparrow$} \\
\midrule
DressCode~\cite{he2024dresscode}~\pub{SIGGRAPH2024}      
& 0.4524 & 0.6803 & 0.8311 \\
ChatGarment~\cite{bian2025chatgarment}~\pub{CVPR2025}  
& 0.4648 & 0.7001 & 0.7946 \\
\rowcolor{gray!20}
Fashion-3DLR~\pub{ours}   & \textbf{0.2743} & \textbf{0.5946} & \textbf{0.9271} \\
\bottomrule
\end{tabular}}
\label{tab:comparisons-try-on}
\end{table}

\begin{table}[htbp]
\centering
\caption{Ablation study on components of Fashion-3DLR.}
\resizebox{0.48\textwidth}{!}{
\begin{tabular}{lcccc}
\toprule
\rowcolor{gray!20}
\textbf{Method} & \textbf{FID$_{\text{CLIP}}$ $\downarrow$} & \textbf{LPIPS $\downarrow$} & \textbf{CLIP-I$_{\text{score}}$ $\uparrow$} \\
\midrule
Ours w/o GFF-DiT  
& 0.6874 & 0.7429 & 0.6983 \\
Ours (full model) 
& \textbf{0.2743} & \textbf{0.5946} & \textbf{0.9271}\\
\bottomrule
\end{tabular}}
\label{tab:Ablation}
\end{table}

\begin{table}[htbp]
\centering
\caption{Comparison on virtual try-on of User-Pref.}
\resizebox{0.48\textwidth}{!}{
\begin{tabular}{lcccc}
\toprule
\textbf{Method}& \textbf{DressCode} & \textbf{ChatGarment}  & \textbf{Fashion-3DLR}   \\
\midrule
\textbf{User-Pref.(\%)$\uparrow$} & 64.41 & 73.25 & \textbf{83.82} \\
\bottomrule
\end{tabular}}
\label{tab:user-try-on}
\end{table}

\subsection{Ablations and Applications}
\myheading{3DGS Physical Experiments.}
We developed our simulation framework upon the MPM to model the physical dynamics of multiple 3DGS-generated garments.
Initially, the generated 3DGS models are scaled, normalized, and positioned within a cubic simulation domain. 
A small subset of kernels located at the top is fixed in a static state to serve as anchor points, while an external impulse is briefly applied to the remaining kernels, enabling natural motion governed by fabric dynamics.
We simulated a variety of material types, including Cotton, Silk, Wool, and Nylon, each represented as a distinct 3DGS physical material. 
The corresponding results are presented in Fig.~\ref{fig:physics}, where the simulation accurately preserves fine-scale geometric details, particularly in the slender band-like structures of bodysuit garments.
As shown in Fig.~\ref{fig:physics}, dynamic demonstrations of 3DGS clothing generated from sketch and oil-brush inputs reveal realistic wave-like deformations and motion trajectories that faithfully capture the intrinsic physical behavior of different fabrics.

\myheading{Virtual Try-on.}
We compare our method with ChatGarment and DressCode, two representative approaches that have shown strong performance in this domain, to evaluate the garment quality and try-on realism achieved by Fashion-3DLR.
Since ChatGarment and DressCode cannot directly take sketch images and texture images as input, we use our proposed GFF-DiT module to process these inputs.
As shown in Fig.~\ref{fig:try-on}, ChatGarment and DressCode generate textures with noticeable color deviations and introduce unrealistic details, such as leather-like dotted patterns. 
In addition, the generated garments exhibit a V-shaped hem that is inconsistent with the flat hem specified in the sketch image, and short sleeves are incorrectly produced.
In contrast, our Fashion-3DLR generates garments whose texture is highly consistent with the reference texture image, showing a more natural horizontal yarn-like distribution. 
The generated torso and sleeves are nearly equal in length, the hem remains flat, and the overall result faithfully preserves the characteristics of both the sketch image and the texture image.
The quantitative results and user study are presented in Tab.~\ref{tab:comparisons-try-on} and Tab.~\ref{tab:user-try-on}.
Our Fashion-3DLR achieves superior performance, demonstrating its strong potential for virtual try-on applications.

\myheading{Ablation Study.}
To further assess our Fashion-3DLR framework, we conducted an ablation study focusing on the GFF-DiT module.
Experimental results indicate that the GFF-DiT component plays a crucial role in understanding and preserving features from semantically diverse image inputs. 
The pre-trained weights obtained from large-scale 2D image data substantially enhance 3D generation quality by mitigating feature inconsistency and preventing generation collapse.
As shown in Tab.~\ref{tab:Ablation}, removing GFF-DiT leads to a marked decline in performance across all evaluation metrics.
The consistent improvements observed in both generation fidelity and condition alignment, validate the pivotal role of GFF-DiT in achieving robust and high-quality 3D garment synthesis.

\section{Conclusion}
We presented Fashion‑3DLR, a controllable 3D garment generation framework that fuses heterogeneous fashion elements into a unified latent space for high‑quality 3D synthesis. 
Through GFF‑DiT, flow‑based latent modeling, and dual 3D Gaussians or mesh decoding, Fashion‑3DLR effectively integrates multi‑modal design cues and produces controllable 3D garments.
Moreover, Fashion‑3DLR supports fully 3D Gaussian‑based physical simulation, enabling material‑dependent dynamic behaviors that remain faithful to the characteristics of diverse fabrics.
Experiments demonstrate superior controllability and generation quality, establishing Fashion‑3DLR as an effective tool for intelligent 3D fashion design.
Although Fashion‑3DLR advances 3D fashion generation, it still faces challenges. 
3D Gaussian garments cannot interact with mesh‑based human bodies, and mesh garments lack sewing‑pattern structures. 
Addressing cross‑representation simulation and pattern‑level modeling will be key directions for future work.

\bibliographystyle{ACM-Reference-Format}
\balance
\bibliography{cvpr}

\end{document}